\pdfoutput=1

\documentclass[11pt]{article}

\usepackage[review]{ACL2023}

\usepackage{times}
\usepackage{latexsym}

\usepackage[T1]{fontenc}

\usepackage[utf8]{inputenc}

\usepackage{microtype}

\usepackage{inconsolata}

\usepackage{graphicx} 
\usepackage{algpseudocode}
\usepackage{algcompatible}
\usepackage{todonotes}
\usepackage{amssymb}
\usepackage{xcolor}
\usepackage{array,multirow}
\usepackage{pifont}
\usepackage{algorithm}
\usepackage{amsmath}
\usepackage{mathtools}
\usepackage{upgreek}
\usepackage{float}
\usepackage[normalem]{ulem}

\title{DyLoRA: Parameter Efficient Tuning of Pre-trained Models using \underline{Dy}namic Search-Free \underline{Lo}w \underline{R}ank \underline{A}daptation }

\author{First Author \\
  Affiliation / Address line 1 \\
  Affiliation / Address line 2 \\
  Affiliation / Address line 3 \\
  \texttt{email@domain} \\\And
  Second Author \\
  Affiliation / Address line 1 \\
  Affiliation / Address line 2 \\
  Affiliation / Address line 3 \\
  \texttt{email@domain} \\}

\begin{document}
\maketitle
\begin{abstract}
With the ever-growing size of pre-trained models (PMs), fine-tuning them has become more expensive and resource-hungry. As a remedy, low-rank adapters (LoRA) keep the main pre-trained weights of the model frozen and just introduce some learnable truncated SVD modules (so-called LoRA blocks) to the model. While LoRA blocks are parameter efficient, they suffer from two major problems: first, the size of these blocks is fixed and cannot be modified after training (for example, if we need to change the rank of LoRA blocks, then we need to re-train them from scratch); second, optimizing their rank requires an exhaustive search and effort. In this work, we introduce a dynamic low-rank adaptation (DyLoRA) technique to address these two problems together. Our DyLoRA method trains LoRA blocks for a range of ranks instead of a single rank by sorting the representation learned by the adapter module at different ranks during training. We evaluate our solution on different natural language understanding (GLUE benchmark) and language generation tasks (E2E, DART and WebNLG) tasks using different pre-trained models such as RoBERTa and GPT with different sizes.  
Our results show that we can train dynamic search-free models with DyLoRA at least $7\times$ faster than LoRA without significantly compromising performance. Moreover, our models can perform consistently well on a much larger range of ranks compared to LoRA.
 \footnote{The source code is publicly available at Anonymized!
 }

 
\end{abstract}

\section{Introduction}

\begin{figure*}[htb!]
    \centering
    \includegraphics[width=1.0\textwidth]{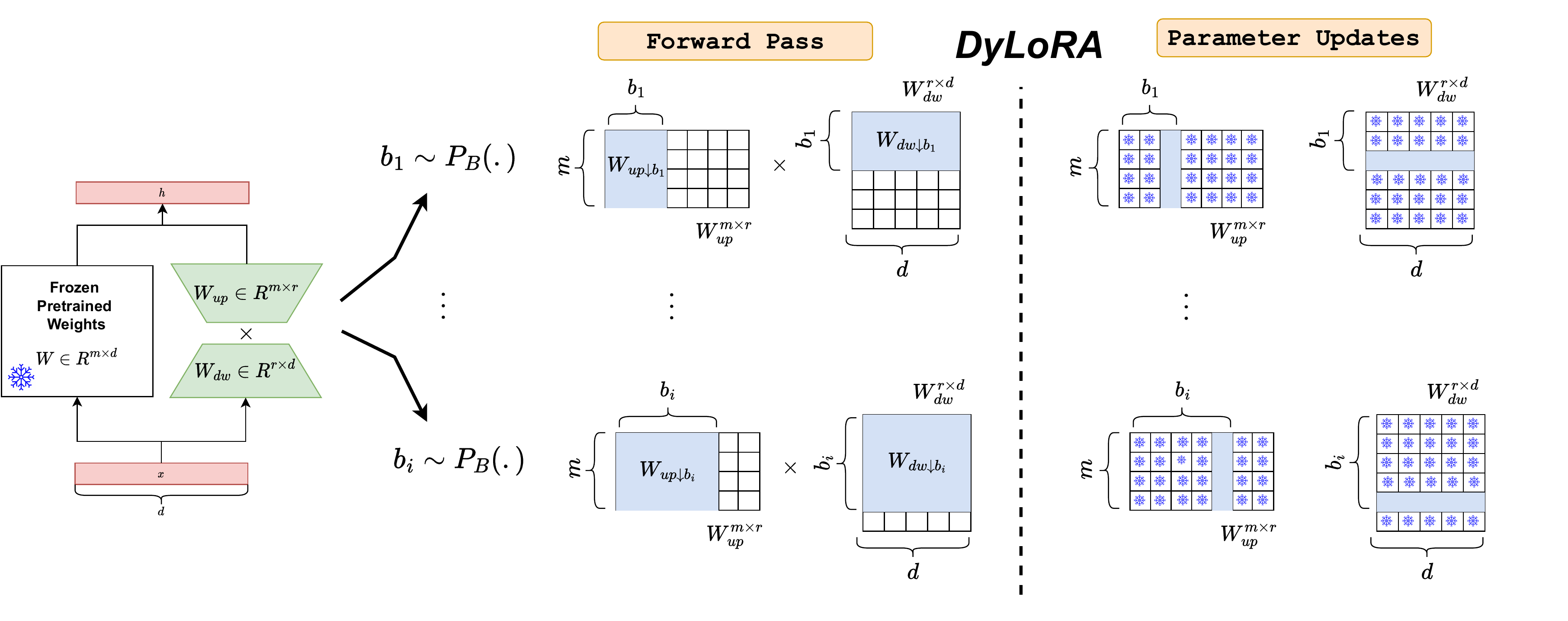}
    \caption{DyLoRA: The overall diagram of our proposed method. In each iteration, we sample from a pre-defined random distribution which will help us to truncate the up-projection and down-projection matrices in the LoRA \cite{hu2021lora} objective.}
    \label{fig:DyLoRA}
\end{figure*}

Pre-training/fine-tuning has become a popular paradigm for solving many tasks in natural language processing (NLP)~\cite{devlin2018bert,liu2019roberta,brown2020language} and Computer Vision~\cite{simonyan2014very, he2016deep, howard2019searching, bochkovskiy2020yolov4, chen2020generative, dosovitskiy2020image}. 
Pre-trained models (PMs) such as pre-trained language models (PLMs)~\cite{devlin2018bert, brown2020language},  and pre-trained visual-language models~\cite{lu2019vilbert, li2019visualbert, su2019vl, xia2021xgpt} have advanced a lot in recent years. 
With the ever-growing size of these pre-trained models, fine-tuning them on downstream tasks becomes more expensive. Moreover, as the ratio of the number of parameters of models with respect to the labeled data increases, the fine-tuning process will be more prone to overfitting~\cite{karimi2021compacter}. 
There are two categories of solutions: first, model compression~\cite{jafari2021annealing, chen2021drone}; second, parameter efficient tuning (PET)~\cite{houlsby2019parameter, karimi2021compacter, mao2021unipelt}. 

There are many different model compression techniques in the literature for Transformer-based models such as matrix factorization~\cite{noach2020compressing, tahaei2021kroneckerbert}, pruning~\cite{wang2019structured}, quantization~\cite{tao2022compression,prato2020fully}, and knowledge distillation~\cite{hinton2015distilling, li2021dynamic, jafari2021annealing,passban2020alp,rashid2021mate}. There are also different types of PET techniques in the literature such as low-rank adapters~\cite{wang2020k, karimi2021compacter,adapter,lora}, and prompt-based techniques~\cite{lester2021power}.     

Although model compression solutions are well-established in recent years in the literature, applying them to large language models can be very costly, because compression techniques usually need to train (or fine-tune) the original large model. 
A case in point is knowledge distillation which relies on fine-tuning a large teacher model or even pre-training the student model as suggested in~\cite{jiao2019tinybert}. Moreover, using compression techniques usually leads to degrading the model performance. PETs can be alternatives to the compression methods, especially when we would like to use the full capacity of the large pre-trained models with light training efforts (such as the \textit{language-model-as-a-service} scenario~\cite{sun2022black}). Among PET techniques, low-rank adapters have received much attention because, in contrast to prompt-tuning techniques, low-rank adapters do not add to the sequence length, get trained faster, and perform better~\cite{karimi2021compacter}. 
Even though there are several low-rank adaptation techniques in the literature, such as Adapter ~\cite{adapter}, Compacter~\cite{karimi2021compacter}, and LoRA~\cite{lora}; they all suffer from two major common problems: first, it is not clear how to select the size of their rank (while their performance is very sensitive to this rank selection); second, their training is static which means that if a low-rank model is trained based on a particular rank size, it will not work well in other rank values (i.e. for any other rank value we need to train a separate model).  

This paper proposes a dynamic low-rank adapter technique (DyLoRA) to address these two problems. Without loss of generality, we focus on LoRA\cite{hu2021lora} and train LoRA blocks for a range of ranks instead of a single rank by sorting out the representation learned at different ranks during training. 
While our model is more flexible, it can outperform LoRA in a much wider range of ranks without adding to the training time. Moreover, our technique does not need extra training for searching across ranks. 
We summarize our contributions in the following:
\begin{itemize}
    \item Dynamic LoRA: On top of LoRA, we developed a new algorithm (DyLoRA) that makes it dynamic at inference time without incurring extra costs.
    \item Search-free LoRA: We demonstrate that by making a negligible compromise in performance, it is possible to avoid the costly search process of choosing the optimal rank for LoRA.
\end{itemize}


\section{Related Work}
This section reviews low-rank adaptation techniques for parameter efficient tuning and potential existing solutions to make these techniques dynamic and search-free.  

It has been shown in~\cite{aghajanyan2020intrinsic} that for classification tasks such as natural language understanding (NLU), PLMs have a low intrinsic dimension. This observation motivates the use of low-rank adapters for parameter efficient tuning. 
There are several low-rank adapters in the literature such as LoRA \citep{lora}, Adapter~\citep{adapter}, Compacter~\citep{karimi2021compacter}, and Parallel Adapter (PA)~\citep{he2021towards}. LoRA is a low-rank up-projection/down-projection transformation without any non-linearity applied in parallel to key and value attention matrices. The main benefit of LoRA is that the adapter module, after training, can be integrated into the original weight matrices of the model, which in turn can lead to a very efficient inference time. Adapters also have a low-rank up-projection/down-projection transformation with an intermediate non-linearity. The Adapter module is applied in series with the feed-forward network (FFN). Having the adaptor module in-line with other blocks in the model can increase the inference time of the model. PA is a faster version of the Adapter, which can be applied in parallel with the FFN block. The compactor is a more memory-efficient version of the Adapter, which deploys the sum of Kronecker products to reconstruct each up-projection and down-projection matrices. All these low-rank adapters suffer from two major issues: first, finding the best rank requires heavy exhaustive training and search; second, the tuned adapter module works well only with a particular rank.

While there have been some efforts in the literature towards dynamic networks such as DynaBERT~\citep{hou2020dynabert} and GradMax \citep{evci2022gradmax}, to the best of our knowledge, this problem for factorized networks and low-rank adapters is still open. DRONE~\cite{chen2021drone} propose a technique for data-aware low-rank model compression however their approach is not search-free, and also, it is not dynamic. DynaBERT introduces a two-stage method to train width and depth-wise dynamic networks. However, DynaBERT requires a fine-tuned teacher model on the task to train its sub-networks which makes it unsuitable for PET techniques.  
GradMax is a technique that gradually adds to the neurons of a network without touching the already trained neurons. But it is unclear how GradMax can be deployed to alleviate the rank-search problem in low-rank adapters. \citet{wang2019structured} propose a structured pruning technique called factorized low-rank pruning (FLOP). FLOP decomposes weight matrices of a network into the sum of rank-1 components, which are regularized during training to gain sparsity. It is worth mentioning that FLOP aims at compressing the main model, and even if it can be used for finding a good rank in the lower-rank representation of full-weight matrices, the final low-rank model will not be dynamic (i.e. it is trained well only for one rank and not a range of ranks, same as LoRA.). 
In this paper, we propose a new methodology for training low-rank modules for multiple ranks simultaneously rather than training a single-rank adapter at a time (without changing the training budget). Inspired by the idea of \textit{nested dropout}~\cite{nested_drop}, we pursue ordering the representations of the bottleneck at the low-rank adapter modules with a new recipe. To the best of our knowledge, it is the first time that the concept of ordering representations has been deployed in training PLMs. 

\section{Background }
\subsection{Nested Dropout}
Inspired by the dropout~\cite{hinton2012improving}, nested drop-out~\cite{nested_drop} is a stochastic regularization technique that targets enforcing ordered representations in training auto-encoders. The nested dropout, adds an implicit bias (which does not exist in dropout) to favor order in training. For example, in dropout, we can randomly drop any nodes or units in the network, but in nested dropout, if we randomly select $k^{\text{th}}$ unit, then we keep all the units indexed from $1$ to $k$ and drop the units with indices larger than $k$. Therefore, nested dropout tends toward accommodating more important information in lower indices while learning representations.

Following the notations of \citep{nested_drop}, nested dropout assumes an auto-encoder mapping of $N$ training examples $\{y_i\}_{i=1}^N \in Y $ , $ Y \subset \mathbb{R}^D$ to their corresponding representations $\{x_i\}_{i=1}^N \in X$, $X \subset \mathbb{R}^K$ using the function $f_\theta:Y\rightarrow X$ with parameters $\theta$; and then decoding these representations using another function $g_\psi: X\rightarrow Y$ with parameters $\psi$ to reconstruct the inputs. The reconstruction loss can be defined as follows:  
\begin{equation}
\begin{split}
        C(\theta, \psi)= \sum_{i=1}^N ||y_i - g_\psi(f_\theta (y_i)) ||^2. 
\end{split}
\end{equation}
Suppose we want to randomly drop some units in our representation vector x. In this regard, we sample a random variable $b\sim p_B(.), b\in \{1, 2, ..., K\}$ from a pre-defined categorical distribution $p_B(.)$ and truncate the functions $f_\theta$ and $g_\psi$ to keep their corresponding units indexed from 1 to $b$ and dropping $b+1$ to $K$ indices. Let's define the b-truncated version of the vector $x$ as $x_{\downarrow b}$ and the b-truncated version of the functions $f_\theta$ and $g_\psi$ as $f_{\theta \downarrow b}$ and $g_{\psi \downarrow b}$ respectively. In this case, the reconstruction loss is redefined for the b-truncated model as follows: 
\begin{equation}
\begin{split}
        &C(\theta, \psi)= \mathbb{E}_{p_B}[C_{\downarrow b}(\theta, \psi)] =  \sum_{b=1}^K p_B(b) C_{\downarrow b}(\theta, \psi)\\
        &\text{where} \\
        &  C_{\downarrow b}(\theta, \psi)= \sum_{i=1}^N ||y_i - g_{\psi \downarrow b} (f_{\theta\downarrow b} (y_i)) ||^2.
\end{split}
\end{equation}
In the final stage, the parameters of this model can be obtained by solving the following optimization problem.    

\begin{equation}
\begin{split}
        (\theta^*, \psi^*) =  \underset{\theta, \psi}{\text{argmin }} C(\theta, \psi).  
\end{split}
\end{equation}
While our work in this paper is inspired by the feature of ordering information suggested in nested dropout, we can distinguish our work from nested dropout in several aspects: 
\begin{enumerate}
    \item  The nested dropout technique is used to add order information to a vector representation; however, we are adding order information to the low-rank matrix decomposition to make it work across a range of ranks instead of a single rank.
    \item Our training algorithm differs from nested dropout in the choice of the distribution function $p_B(.)$, and we propose a more efficient  individual loss for each truncated matrix compared to the linear summation loss in nested dropout. 
\end{enumerate}

\subsection{LoRA: Low-rank Adapters}
In LoRA~\cite{hu2021lora}, some pre-trained weights of dense layers of PLMs are summed with parallel linear low-rank adapter modules. During fine-tuning, the original pre-trained weights are kept frozen; LoRA modules can be updated instead. For example, let's assume that $W_0 \in \mathbb{R}^{m\times d}$ is a pre-trained weight matrix in the network which is accompanied by a LoRA module $\Delta W = W_{up}W_{dw}$ where $W_{up}\in \mathbb{R}^{m\times r}$, $W_{dw}\in \mathbb{R}^{r\times d}$, and $r \ll min(m,d)$. Then, the output of this layer can be obtained as 
\begin{equation}
\begin{split}
    h = W_0x + \Delta W x= W_0 x+ \frac{\alpha}{r}W_{up}W_{dw}x.
\end{split}
\end{equation}
Bear in mind that the $W_{up}$ matrix is initialized as a zero matrix, and the $W_{dw}$ matrix is initialized as a zero-mean Gaussian distribution where $\alpha$ is a constant scale hyper-parameter.

In LoRA, the rank $r$ is a hyperparameter that should be tuned for each task. 
Moreover, LoRA is a \textit{static} low-rank adapter that works only with a particular size of $r$, which is trained based on.

\section{Our Method: DyLoRA}


\begin{table*}[hbt!]
\centering
\begin{tabular}{lcccccc}
\hline
\multicolumn{7}{c}{Model: RoBERTa-Base} \\
\hline
\textbf{Task} & \textbf{Rank=1} & \textbf{Rank=2} & \textbf{Rank=4} & \textbf{Rank=8} & \textbf{Rank=16} & \textbf{Rank=32}\\
\hline
\textbf{QQP} (Accuracy) & \underline{89.14} & 89.96 & 90.33 & 90.69 & 90.95 & \textbf{91.02} \\
\textbf{SST-2} (Accuracy) & \underline{93.58} & 94.15 & 94.38 & \textbf{94.84} & 94.27 & 94.5 \\
\textbf{MRPC} (Accuracy) & 87.25 & 87.75 & 88.24 & 87.25 & \underline{86.76} & \textbf{89.22} \\
\textbf{CoLA} (Mathews) & 61.84 & \underline{57.78} & 61.57 & \textbf{63.81} & 63.07 & 62.82\\
\hline
\end{tabular}
\caption{\label{rank-effect}
The effect of the rank of the low-rank adaptation matrix over the performance of the model. 
In this experiment, all the other hyperparameters are fixed, and we only changed the rank of the LoRA model. In this search space, \underline{Underline} shows the minimum performance rank, and the \textbf{bold} number shows the maximum performance rank. 
}
\end{table*}

In this section, we introduce our solution to get dynamic low-rank adapters that can be trained and deployed well on a range of ranks instead of a single particular rank (with a fixed training budget). This flexibility can free us from searching for the best ranks by training the model multiple times.

Without loss of generality, we explain our solution on top of LoRA as one of the prominent low-rank adapter techniques in the literature. In each LoRA module, we have an up-projection ($W_{up}\in \mathbb{R}^{m\times r}$) and a down-projection matrix ($W_{dw}\in \mathbb{R}^{r\times d}$).   
Let's assume that we would like to train the LoRA module to operate in the range of $r \in \text{Range}[r_{min}, r_{max}]$ where $r_{min}$ and $r_{max}$ can be treated as new hyper-parameters. 
To make the LoRA module work in a range of ranks instead of a single rank, we need to ensure that increasing or decreasing the rank will not significantly hamper the model's performance. One way to implement such behavior would be by sorting the information content of different ranks in the training process of LoRA modules.   
In this regards, at each training step, we sample $b\sim p_B(.), b\in\{r_{min}, r_{min}+1, ..., r_{max}\}$ form a pre-defined categorical distribution (which has a support in $\text{Range}[r_{min}, r_{max}]$) and truncate $W_{dw}$ and $W_{up}$ matrices accordingly. 
\begin{equation}
\begin{split}
 & W_{dw \downarrow b} = W_{dw}[1:b,:] \\
 & W_{up \downarrow b} = W_{up}[:,1:b] \\
 \end{split}
\end{equation}
$W_{dw \downarrow b}$ and $W_{up \downarrow b}$ are b-truncated versions of $W_{dw}$ and $W_{up}$ respectively (see Fig.~\ref{fig:DyLoRA} for the visualization). 
Moreover, let's define $W_{dw}^b$ as the $b^\text{th}$ row of $W_{dw}$; $W_{up}^b$ corresponds to the $b^\text{th}$ column of $W_{up}$. 
\begin{equation}
\begin{split}
 & W_{dw}^b = W_{dw}[b,:] \\
 & W_{up}^b = W_{up}[:,b] \\
\end{split}
\end{equation}
Then, the forward pass of this truncated LoRA module during training will be calculated as following: 
\begin{equation}
\begin{split}
  & h = W_0x+ \frac{\alpha}{b} W_{up \downarrow b} W_{dw \downarrow b} x 
\end{split}
\label{eq: LoRA}
\end{equation}

For simplicity, let's assume that we have only one LoRA module in the network (the one which is described in~Eq.~\ref{eq: LoRA}). Let's first consider the regular static loss function ($\mathcal{L^S}$) of the network $f(x;W_{dw},W_{up})$ with $W_{dw}$ and $W_{up}$ tunable parameters for $N$ given input-output pairs  $(\textbf{x},\textbf{y}) = (x_i,y_i)_{i=1}^N$:  
\begin{equation}
\begin{split}
  &  \underset{W_{dw}, W_{up}}{\text{min}} \mathcal{L^S} (\textbf{x},\textbf{y};W_{dw},W_{up}) \triangleq \\ 
  &  \sum_{i=1}^N l (f(x_i;W_{dw},W_{up}),y_i). 
\end{split}
\end{equation}
where $l(f,\textbf{y})$ is a loss function that measures the divergence of network predictions compared with the target labels. Then, let's extend the training loss to make the network dynamic considering the b-truncation process. We can define our dynamic loss function $\mathcal{L^{DY}}$ as follows.
\begin{equation}\label{eq:lossFunction}
\begin{split}
  &  \mathcal{L}_{\downarrow b}^{\mathcal{DY}} = \sum_{i=1}^N l (f(x_i;W_{dw \downarrow b},W_{up \downarrow b}),y_i). 
\end{split}
\end{equation}
Bear in mind that, our loss function has a major difference from the nested dropout loss, which makes it more efficient. The nested dropout loss is in the form of $\sum_{b=r_{min}}^{r_{max}} p_B(b) \mathcal{L}_{\downarrow b}^{\mathcal{DY}} (\textbf{x},\textbf{y};W_{dw\downarrow b},W_{up\downarrow b}) $ which requires to sum the loss over the entire possible range of ranks and it is computationally expensive. To overcome this computational restriction, we replace it by optimizing the model parameters for each target rank individually at each time step. We show that this scheme quite works well. 

The other difference with nested dropout is that in the parameter update phase, we add a new mode (so-called \textit{frozen}) as a hyper-parameter to our training. This new mode suggests to only update the $b^\text{th}$ corresponding row and column sampled in the truncation phase (i.e. a single row or column will be updated at a time to prevent the learning parameters from being forgotten at previous time steps.). This approach can enhance the efficiency of our algorithm even more. Table~\ref{ablation-study} shows the impact of only updating "b" versus updating the columns and rows from 1 to $b$.

\begin{equation}
\begin{split}
  &  W_{dw}^b  \leftarrow W_{dw}^b - \eta \nabla_{W_{dw}^b} \mathcal{L}_{\downarrow b}^{\mathcal{DY}} \\
  &  W_{up}^b  \leftarrow W_{up}^b - \eta \nabla_{W_{up}^b} \mathcal{L}_{\downarrow b}^{\mathcal{DY}} 
  \end{split}
\end{equation}

The summary of our technique is described in Algorithm~\ref{alg} in Appendix \ref{ap:algorithm}.

\section{Experiments}

\begin{table*}[hbt!]
\centering
\resizebox{\textwidth}{!}{
\begin{tabular}{lccccccccc}
 & \textbf{Accuracy} & \textbf{Accuracy} & \textbf{F1} & \textbf{Mathews} & \textbf{Accuracy} & \textbf{Accuracy} & \textbf{Accuracy} & \textbf{Pearson} & \\
\hline
\textbf{Model} & \textbf{MNLI} & \textbf{SST-2} & \textbf{MRPC} & \textbf{CoLA} & \textbf{QNLI} & \textbf{QQP} & \textbf{RTE} & \textbf{STS-B} & \textbf{Avg}\\
\hline
\multicolumn{10}{c}{Rank = 1} \\ 
\hline
LoRA & $34.60_{\pm3.69}$ &	$69.61_{\pm7.99}$&	$83.47_{\pm3.90}$& $25.57_{\pm9.71}$&	$53.00_{\pm2.95}$&	$44.30_{\pm7.50}$&	$57.55_{\pm5.51}$&	$76.07_{\pm6.06}$&	54.90 \\
\textbf{DyLoRA} & $85.59_{\pm0.07}$ &	$93.23_{\pm0.63}$&	$91.58_{\pm0.69}$& $57.93_{\pm2.12}$&	$91.95_{\pm0.14}$&	$88.37_{\pm0.15}$&	$74.80_{\pm1.48}$&	$90.30_{\pm0.13}$&	\textbf{84.22} \\
\hline
\multicolumn{10}{c}{Rank = 2} \\ 
\hline
LoRA &  $40.53_{\pm6.17}$ &	$82.75_{\pm5.08}$&	$88.00_{\pm1.81}$& $43.30_{\pm4.67}$&	$63.42_{\pm2.99}$&	$59.21_{\pm6.13}$&	$68.88_{\pm1.26}$&	$85.51_{\pm1.94}$&	66.45 \\
\textbf{DyLoRA} & $86.02_{\pm0.06}$ &	$93.81_{\pm0.30}$&	$91.66_{\pm0.46}$& $59.91_{\pm1.88}$&	$92.39_{\pm0.25}$&	$89.33_{\pm0.05}$&	$76.03_{\pm1.61}$&	$90.60_{\pm0.09}$&	\textbf{84.97} \\
\hline
\multicolumn{10}{c}{Rank = 3} \\ 
\hline
LoRA &  $58.95_{\pm6.02}$ &	$90.00_{\pm1.27}$&	$89.66_{\pm1.25}$& $56.78_{\pm1.88}$&	$79.26_{\pm4.80}$&	$72.58_{\pm4.09}$&	$72.49_{\pm2.30}$&	$88.80_{\pm0.29}$&	76.07 \\
\textbf{DyLoRA} & $86.70_{\pm0.09}$ &	$94.11_{\pm0.33}$&	$91.56_{\pm0.86}$& $60.97_{\pm2.01}$&	$92.77_{\pm0.21}$&	$89.76_{\pm0.07}$&	$77.11_{\pm2.97}$&	$90.69_{\pm0.14}$&	\textbf{85.46} \\
\hline

\multicolumn{10}{c}{The full table can be found in the Appendix \ref{ap:roberta-base-experiments}} \\ 
\hline
\multicolumn{10}{c}{Rank = 7} \\ 
\hline
LoRA &  $85.44_{\pm0.78}$ &	$93.62_{\pm0.35}$&	$91.27_{\pm0.73}$& $62.19_{\pm2.66}$&	$91.88_{\pm0.23}$&	$89.51_{\pm0.30}$&	$75.52_{\pm1.41}$&	$90.35_{\pm0.24}$&	84.97 \\
\textbf{DyLoRA} & $86.82_{\pm0.10}$ & $94.27_{\pm0.33}$ &	$91.38_{\pm0.59}$ & $59.51_{\pm1.75}$ & $92.99_{\pm0.26}$ & $90.04_{\pm0.06}$ & $77.91_{\pm1.58}$ & $91.07_{\pm0.19}$ &	\textbf{85.50} \\

\hline
\multicolumn{10}{c}{Rank = 8} \\ 
\hline
LoRA &  $86.82_{\pm0.18}$ & $94.01_{\pm0.30}$ & $91.48_{\pm0.73}$ & $62.08_{\pm1.37}$ & $92.39_{\pm0.39}$ & $90.42_{\pm0.02}$ &	$74.51_{\pm0.41}$ & $90.48_{\pm0.24}$ &	85.27 \\
\textbf{DyLoRA} & $86.76_{\pm0.13}$ & $94.36_{\pm0.38}$ & $91.38_{\pm0.83}$ &	$59.51_{\pm1.84}$ & $93.00_{\pm0.32}$ &	$89.91_{\pm0.08}$ & $77.55_{\pm0.59}$ &	$91.05_{\pm0.19}$ & \textbf{85.44} \\
\hline
\multicolumn{10}{c}{Best (Rank)} \\ 
\hline
LoRA & 87.03(8) & 94.50(6) & 92.25(7) & \textbf{66.05}(7) & 92.81(8) & \textbf{90.45}(8) & 77.98(6) & 90.87(8) & 86.49 \\
\textbf{DyLoRA} & \textbf{87.17(6)} & \textbf{94.72}(7) & \textbf{92.79}(8) &	63.32\textbf{(3)} & \textbf{93.56}(8) & 90.17\textbf{(6)} & \textbf{80.14(4)} & \textbf{91.36}(7) & \textbf{86.66} \\
\hline
\multicolumn{10}{c}{Full Rank} \\ 
\hline
Fine Tune$^*$ & \textbf{87.6} & \textbf{94.8} & 90.2 & 63.6 & 92.8 &\textbf{91.9} &78.7 &91.2 & 86.4 \\
\hline
\end{tabular}
}
\caption{\label{fixed-budget}
In this table, the task is to find a low-rank adaptation matrix that works with different ranks at inference time given a fixed budget (training time). In all experiments, the pre-trained model is RoBERTa-Base \cite{liu2019roberta}. Following LoRA \cite{hu2021lora}, the last row, which is indicated by *, reports the results of "Fine Tune" from the original \cite{liu2019roberta} paper. 
}
\end{table*}

\begin{table}[hbt!]
\centering
\small
\begin{tabular}{lccc}
& & Accuracy & Accuracy \\
\hline
\textbf{Model} & \textbf{Time} & \textbf{SST-2} ($r$) & \textbf{MRPC} ($r$)  \\
\hline
\multicolumn{4}{c}{Maximum Rank: $r_{max}=64$} \\ 
\hline
LoRA (Search) &  $7x$ & \textbf{95.3}(64)	 &	89.71(64)   \\
\textbf{DyLoRA} &  $\textbf{1x}$ & 94.38\textbf{(7)} &	\textbf{89.95(34)}  \\
\hline
\multicolumn{4}{c}{Maximum Rank: $r_{max}=32$} \\ 
\hline
LoRA (Search) &  $6x$ & \textbf{94.84}(32) &	88.73(16)   \\
\textbf{DyLoRA} &  $\textbf{1x}$ & 94.38\textbf{(7)} &	\textbf{89.71(5)}  \\





\hline
\end{tabular}
\caption{
 In this table, the search space of rank is larger compared to the previous experiment and the goal is to find the most optimal rank for the low-rank adaptation of a pre-rained RoBERTa-Base. The lower the rank the better, and the higher the performance is the better. 
}
\label{search-experiment}
\end{table}



\begin{table*}[hbt!]
\centering
\resizebox{\textwidth}{!}{
\begin{tabular}{lccccccc}
& & Accuracy & F1 & Accuracy & Pearson & \\
\hline
\textbf{Model} (Rank) & \textbf{Trainable Params} & \textbf{SST-2} & \textbf{MRPC} & \textbf{QNLI} & \textbf{STS-B} & \textbf{AVERAGE}  \\
\hline
Fine Tune$^*$ & 125M & \textbf{94.8} & 90.2 & 92.8 & \textbf{91.2} & 92.25 \\
FLOP$^*$ &  80M & 92.09 &	88.61  & 89.05 & 88.18 & 89.48  \\
LoRA (1) & \textbf{0.628M} & 93.58 & 91.93 & 91.98 & 90.85 &  \\
\hline
\multicolumn{7}{c}{Maximum Rank: $r_{max}=8$} \\ 
\hline
\textbf{DyLoRA} (1)  & \textbf{0.628M} & $93.23_{\pm0.63}$ & $91.58_{\pm0.69}$ &  $91.95_{\pm0.14}$ & $90.30_{\pm0.13}$ & 91.77 \\
\textbf{DyLoRA} (8)  & 0.887M & $94.36_{\pm0.38}$ & $91.38_{\pm0.83}$ & $93.00_{\pm0.32}$ & $91.05_{\pm0.19}$ & \textbf{92.45} \\
\hline
\end{tabular}
}
\caption{
This table compares DyLoRA with compression-based algorithms. 
As indicated by *, we reported "Fine Tune" and FLOP from their original papers, \cite{liu2019roberta} and \cite{wang2019structured}. 
To the best of our knowledge, experiments were conducted under the same experimental setting. We count all the trainable parameters including classifier, unlike LoRA paper \cite{hu2021lora} which they count only LoRA specific parameters. 
}
\label{regularization-prunning}
\end{table*}

\begin{table*}[hbt!]
\centering
\resizebox{\textwidth}{!}{
\begin{tabular}{l|c|ccccccc}
\hline
\multicolumn{9}{c}{Maximum Rank: $r_{max}=8$} \\ 
\hline
& & Accuracy & F1 & Mathews & Accuracy & Accuracy & Pearson & \\
\multirow{1}{*}{$b \sim P_B$: \textit{Distribution}} & \multirow{1}{*}{\textit{Updated Parameters}} & \textbf{SST-2} & \textbf{MRPC} & \textbf{CoLA}  & \textbf{QNLI} & \textbf{RTE}  & \textbf{STS-B} & \textbf{AVERAGE} \\\cline{3-9}
& & \multicolumn{7}{c}{\textbf{Rank=8}} \\ 
\hline
\multirow{2}{*}{\textbf{Geometric (p=0.15)}} & $W_{dw \downarrow b}$,$W_{up \downarrow b}$ & $93.97_{\pm0.33}$  & $90.84_{\pm1.15}$& $58.95_{\pm1.95}$& $92.74_{\pm0.13}$& $74.80_{\pm0.90}$ & $90.66_{\pm0.15}$ & 83.66 \\
& $W_{dw}^b$,$W_{up}^b$ & $93.60_{\pm0.24}$ & $90.50_{\pm0.42}$ & $58.19_{\pm1.17}$ & $92.26_{\pm0.12}$ & $71.91_{\pm1.74}$ & $90.20_{\pm0.36}$ & 82.78 \\
\hline
\multirow{2}{*}{\textbf{Uniform}} & $W_{dw \downarrow b}$,$W_{up \downarrow b}$ & $94.36_{\pm0.38}$ & $91.38_{\pm0.83}$& $59.51_{\pm1.84}$& $93.00_{\pm0.32}$& $77.55_{\pm0.59}$ & $91.05_{\pm0.19}$ & \textbf{84.47} \\
& $W_{dw}^b$,$W_{up}^b$ & $93.69_{\pm0.41}$ & $91.19_{\pm0.79}$ & $58.52_{\pm0.95}$ & $92.47_{\pm0.18}$ & $73.29_{\pm2.49}$ & $90.68_{\pm0.14}$ & 83.31\\
\hline
& & \multicolumn{7}{c}{\textbf{Rank=1}} \\ 
\hline
\multirow{2}{*}{\textbf{Geometric (p=0.15)}} & $W_{dw \downarrow b}$,$W_{up \downarrow b}$ & $93.53_{\pm0.47}$ & $91.36_{\pm0.72}$& $59.43_{\pm1.12}$ & $92.24_{\pm0.08}$& $73.65_{\pm3.55}$& $90.33_{\pm0.14}$ & \textbf{83.42}  \\
& $W_{dw}^b$,$W_{up}^b$ & $93.58_{\pm0.26}$ & $90.81_{\pm0.83}$ & $58.55_{\pm1.13}$ & $92.27_{\pm0.28}$ & $68.52_{\pm11.88}$ & $90.60_{\pm0.31}$ & 82.39 \\
\hline
\multirow{2}{*}{\textbf{Uniform}} & $W_{dw \downarrow b}$,$W_{up \downarrow b}$ & $93.23_{\pm0.63}$ & $91.58_{\pm0.69}$ & $57.93_{\pm2.12}$ & $91.95_{\pm0.14}$ & $74.80_{\pm1.48}$ & $90.30_{\pm0.13}$ & 83.30 \\
& $W_{dw}^b$,$W_{up}^b$ & $93.51_{\pm0.49}$ & $90.75_{\pm0.70}$ & $56.95_{\pm1.54}$ & $91.70_{\pm0.28}$ & $66.79_{\pm8.54}$ & $89.95_{\pm0.24}$ & 81.61 \\

\hline
\end{tabular}
}
\caption{\label{ablation-study}
Ablation Study -  In this experiment, our goal is to demonstrate how the introduced distribution can affect the performance of DyLoRA. 
}
\end{table*}

In this section, we describe the experiments used to evaluate our DyLoRA model on both natural language understanding (NLU) and natural language generation (NLG) tasks. To be fair with the original LoRA method, we try to keep the setting of our experiments similar to the LoRA paper~\cite{hu2021lora}. 
Therefore similarly, we chose the pretrained RoBERTa \cite{liu2019roberta} base model as the backbone of the LoRA and DyLoRA experiments for the GLUE benchmark, and GPT-Medium for the NLG tasks. For our experiments, we did not use any hyper-parameter tuning, nor did we search the validation epochs, nor did we use MLNI trick (use the MLNI checkpoint instead of the pre-trained weights) to enhance the model's performance. More details about the hyperparameters is available in Table \ref{hyperparamters-table} in Appendix \ref{ap:hyperparameters}. In total, we conducted more than 200 experiments and evaluated more than 1600 models, details of which can be found in the attachments.

\subsection{Baselines}
\begin{itemize}
    \item \textbf{Fine Tune}: To show a relative upper bound for the performance of our proposed method, we fine-tuned all the parameters in the model. Even though we have a large number of trainable parameters, this can help us better understand how higher-rank models perform.
    
    \item \textbf{LoRA}: As a baseline to DyLoRA, we employed the original LoRA model with their tuned hyperparameters \cite{hu2021lora}. As a result, most of the experiments have been conducted in a favorable manner for LoRA. 
    \item \textbf{FLOP}: Due to its flexibility, \textbf{F}actorized \textbf{Lo}w Rank \textbf{P}runing (FLOP) \cite{wang2019structured} can be applied to any matrix multiplication and, therefore, can be used to avoid the search in our problem. However, this baseline lacks the dynamic properties of DyLoRA. We used it to show regularization-based techniques' performance and pros and cons.
\end{itemize}

\subsection{LoRA rank selection problem} 

There is no clear guidance on how to determine the rank for the LoRA algorithm. It is evident in the LoRA paper \cite{hu2021lora} that the performance of models varies a lot with different ranks (e.g. check Tables 15, and 18 in the LoRA paper), and does not indicate any clear trend. We also observe the same problem in the GLUE benchmark. We may argue that theoretically, the rank with the best performance is always the highest. High ranks, however, introduce additional parameters into the adaptive process and this might be undesirable. In practice, as demonstrated in Table \ref{rank-effect}, the most effective rank differs depending on the task. For example, based on the MRPC results, the rank with the lowest performance is 16 while the rank with the highest performance is 32. This is different from SST-2, in which rank 1 is the least performing rank and rank 8 is the most effective rank. Many factors can contribute to this difference, including but not limited to the size of the dataset, hyperparameter selections, hardware configurations and the optimization.

\subsection{Dynamic low rank adaptation}

For example, suppose we have a neural network that we wish to deploy on various devices with different configurations. The use of higher ranks may pose a problem for very sensitive devices as they have a greater number of parameters. Therefore, we must either train several models with different configurations or find the most optimal rank. The cost associated with this is significant, as even in the setting of LoRA, we are required to find the best rank for each task and each device. Using DyLoRA, however, one needs to train one model per task and, as our method is adaptive at inference time, we can deploy it according to our needs. In Table \ref{fixed-budget}, we demonstrate the dynamic properties of DyLoRA. In order to ensure a fair comparison, all LoRA and DyLoRA models in this table have the same model size,  we used the same code and evaluation process, and all models were trained to the same extent. In LoRA, we lose performance when performing inferences for the lower ranks. This occurs because the model has been trained only for rank 8 during training. In DyLoRA, we preserve a high level of performance for lower ranks while competing well with LoRA on rank 8.

\begin{table*}[hbt!]
\centering
\resizebox{\textwidth}{!}{
\begin{tabular}{lcc|cccccc}
\hline
\textbf{Model} (Method) & \textbf{Updated Params} & \textbf{Trainable Params} & \multicolumn{5}{c}{E2E NLG Challenge} \\ 
& & & \textbf{BLEU} & \textbf{NIST} & \textbf{MET} & \textbf{ROUGE-L} & \textbf{CIDEr} \\
\hline
\multicolumn{8}{c}{Rank=1} \\
\hline
GPT-2 M (LoRA) & & 0.09M & 3.4 & 1.2 & 9.2 & 18.8 & 0.1 \\
\textbf{GPT-2 M (DyLoRA)} & $W_{dw}^b$,$W_{up}^b$ & 0.09M & $67.9_{\pm0.2}$ & $8.6_{\pm0.1}$ & $44.9_{\pm0.4}$ & $69.1_{\pm0.3}$  & $2.4_{\pm0.0}$ \\
\hline

\multicolumn{8}{c}{Rank=2} \\
\hline
GPT-2 M (LoRA) & & 0.19M & 47.0 & 6.4 & 34.2 & 56.1 & 1.3 \\
\textbf{GPT-2 M (DyLoRA)} & $W_{dw}^b$,$W_{up}^b$  & 0.19M & $68.8_{\pm0.5}$ & $8.7_{\pm0.02}$ & $45.2_{\pm0.2}$ & $69.8_{\pm0.3}$  & $2.4_{\pm0.0}$ \\
\hline

\multicolumn{8}{c}{The full table can be found in the Appendix \ref{ap:gpt-experiments}} \\


\hline
\multicolumn{8}{c}{Rank=4} \\
\hline
GPT-2 M (LoRA) & & 0.39M & 69.9 & 8.8 & 46.8 & 72.1 & 2.5 \\
\textbf{GPT-2 M (DyLoRA)} & $W_{dw}^b$,$W_{up}^b$  & 0.39M & $68.4_{\pm0.41}$ & $8.7_{\pm0.02}$ & $45.5_{\pm0.6}$ & $69.9_{\pm0.5}$  & $2.4_{\pm0.0}$ \\
\hline
\end{tabular}
}
\caption{
For all metrics, higher is better. Rows with * have been reported based on the LoRA paper. Unlike \cite{hu2021lora}, we included the classifier number of parameters in our trainable parameters count. 
}
\label{gpt-experiments-e2e}
\end{table*}

\subsection{Search-free low rank adaptation} 

The process of selecting a particular rank can be an expensive one as previously mentioned. In Table \ref{search-experiment}, we present an experiment that illustrates the costs associated with such a search for LoRA and DyLoRA. As an example, if one naively wanted to search the entire range of ranks (for example, 64 in the experiment), then they would have to train and evaluate 64 distinct models in order to determine the proper rank. It becomes even more expensive if you search the entire rank space. In the case of uniform search, this cost is less, yet still more expensive (7 times in the experiment) than our proposed method. Therefore, for LoRA (Search), we ran experiments for ranks={1,2,4,8,16,32,64} and we reported the best results. The results demonstrate that our proposed method performs competitively at a much lower cost.


\begin{table}[hbt!]
\centering
\begin{tabular}{lcc}
\hline
\multicolumn{3}{c}{Maximum Rank: $r_{max}=8$} \\ 
\hline
\textbf{Loss} & \textbf{Training Time} & \textbf{CoLA}  \\
\hline
\textbf{$\mathcal{L}_{\downarrow b}^{\mathcal{DY}}$} & \textbf{645.82s} & 52.64  \\

\textbf{$\sum_{} p_B(b) \mathcal{L}_{\downarrow b}^{\mathcal{DY}}$} & 1175.69s & \textbf{54.12}  \\
\hline
\end{tabular}

\caption{
This experiment shows the impact of choosing individual loss vs. summation loss functions on our training. The average performance across all possible ranks (1,2,...,8) is reported. For summation loss to be computationally more feasible, smaller epochs were chosen. A total of seven GPUs were used.
}
\label{fig:sumIndividual}
\end{table}

\subsection{Regularization and pruning} 

An alternative method of avoiding the search problem is using regularization/pruning techniques to determine the intrinsic rank of the weight matrix. In this way, we can reduce the number of parameters of the original matrices; however, we will not have a dynamic model during inference. To illustrate the difference between such methods and DyLoRA, we reported the performance of one of these models, FLOP ~\cite{wang2019structured}, in Table \ref{regularization-prunning}. FLOP utilizes low-rank factorization to create new matrices representing the original weight matrix. Thus, they will have fewer total parameters but require more trainable parameters to reach a comparable performance to DyLoRA. 

\subsection{Generative Tasks}
In this experiment, we evaluat the performance of our model on different natural language generation (NLG) tasks such as the E2E NLG Challenge~\cite{novikova2017e2e}, DART~\cite{nan2020dart} and WebNLG~\cite{gardent2017webnlg}. The results of the E2E task are shown in Table \ref{gpt-experiments-e2e} and due to the space limit, the results of the other two tasks are demonstrated in Appendix \ref{ap:gpt-experiments}. The generative tasks demonstrate a similar pattern as the NLU task, showing that our model is able to work well at wider range of ranks compared with LoRA. 

\subsection{Ablation study} 

In this subsection, we investigate the impact of two design choices in DyLoRA: first, the new distribution $P_B$ hyper-parameter in our technique; second, the impact of updating $W_{dw}^b$ and $W_{up}^b$ parameters instead of the entire $W_{dw \downarrow b}$ and $W_{up \downarrow b}$. The distribution $P_B$ changes the relative importance of the different ranks during the training process.  
To examine the impact of the chosen distribution on DyLoRA's performance, we used two distributions, geometric and uniform. 
As shown in Table \ref{ablation-study}, the geometric distribution, provides a much better method for optimizing the lower ranks, since it pays much more attention to the lower ranks during training, and uniform distribution will give better performance over all ranks. We chose to use uniform distribution in most of our experiments to avoid adding another hyperparameter which is a requirement of the geometric distribution. Moreover, we demonstrate that it is possible to ensure that the optimization of rank $b$ will not negatively affect the performance of the lower ranks ($1$ to $b-1$), while performing reasonably well. As mentioned, this can be accomplished by only updating the unique parameters associated with rank $r$ that do not overlap with lower ranks. 

In addition, in Table \ref{fig:sumIndividual}, we demonstrate the result of using our individual loss (Eq.~\ref{eq:lossFunction}) vs. the nested dropout original objective function in an equal setting. As shown, our proposed objective function is both effective and efficient. Furthermore, it is important to note that the summation loss is not scalable when many ranks are involved. We also discussed the time complexity of LoRA and DyLoRA in Appendix \ref{ap:time_complexity}.

\section{Conclusion}

In this paper, we presented our solution DyLoRA to address two problems in low-rank adapters regarding rank selection and making them dynamic. We showed that DyLoRA can select the rank without requiring multiple re-training and is able to make LoRA dynamic at inference time. As a result, we can avoid the process of searching for the most optimal ranks for many real-life scenarios. It has been demonstrated that DyLoRA performance is comparable with LoRA, yet we can support a wider range of ranks without adding additional time and effort.



\section*{Limitations}

According to LoRA \cite{hu2021lora}, a proper choice of the scalar can improve the results. In order to determine what is the best choice, further investigation is required. Despite our demonstration that uniform distribution can be as effective as specific geometric distribution, further investigation is necessary to evaluate the effect of different distributions on different downstream tasks. As shown in this paper, our algorithm works over a wide range of ranks, but further research is needed to understand the impact of choosing a particular range.

\bibliography{anthology}

\begin{thebibliography}{43}
\expandafter\ifx\csname natexlab\endcsname\relax\def\natexlab#1{#1}\fi

\bibitem[{Aghajanyan et~al.(2020)Aghajanyan, Zettlemoyer, and
  Gupta}]{aghajanyan2020intrinsic}
Armen Aghajanyan, Luke Zettlemoyer, and Sonal Gupta. 2020.
\newblock Intrinsic dimensionality explains the effectiveness of language model
  fine-tuning.
\newblock \emph{arXiv preprint arXiv:2012.13255}.

\bibitem[{Bochkovskiy et~al.(2020)Bochkovskiy, Wang, and
  Liao}]{bochkovskiy2020yolov4}
Alexey Bochkovskiy, Chien-Yao Wang, and Hong-Yuan~Mark Liao. 2020.
\newblock Yolov4: Optimal speed and accuracy of object detection.
\newblock \emph{arXiv preprint arXiv:2004.10934}.

\bibitem[{Brown et~al.(2020)Brown, Mann, Ryder, Subbiah, Kaplan, Dhariwal,
  Neelakantan, Shyam, Sastry, Askell et~al.}]{brown2020language}
Tom Brown, Benjamin Mann, Nick Ryder, Melanie Subbiah, Jared~D Kaplan, Prafulla
  Dhariwal, Arvind Neelakantan, Pranav Shyam, Girish Sastry, Amanda Askell,
  et~al. 2020.
\newblock Language models are few-shot learners.
\newblock \emph{Advances in neural information processing systems},
  33:1877--1901.

\bibitem[{Chen et~al.(2020)Chen, Radford, Child, Wu, Jun, Luan, and
  Sutskever}]{chen2020generative}
Mark Chen, Alec Radford, Rewon Child, Jeffrey Wu, Heewoo Jun, David Luan, and
  Ilya Sutskever. 2020.
\newblock Generative pretraining from pixels.
\newblock In \emph{International conference on machine learning}, pages
  1691--1703. PMLR.

\bibitem[{Chen et~al.(2021)Chen, Yu, Dhillon, and Hsieh}]{chen2021drone}
Patrick Chen, Hsiang-Fu Yu, Inderjit Dhillon, and Cho-Jui Hsieh. 2021.
\newblock Drone: Data-aware low-rank compression for large nlp models.
\newblock \emph{Advances in neural information processing systems},
  34:29321--29334.

\bibitem[{Devlin et~al.(2018)Devlin, Chang, Lee, and
  Toutanova}]{devlin2018bert}
Jacob Devlin, Ming-Wei Chang, Kenton Lee, and Kristina Toutanova. 2018.
\newblock Bert: Pre-training of deep bidirectional transformers for language
  understanding.
\newblock \emph{arXiv preprint arXiv:1810.04805}.

\bibitem[{Dosovitskiy et~al.(2020)Dosovitskiy, Beyer, Kolesnikov, Weissenborn,
  Zhai, Unterthiner, Dehghani, Minderer, Heigold, Gelly
  et~al.}]{dosovitskiy2020image}
Alexey Dosovitskiy, Lucas Beyer, Alexander Kolesnikov, Dirk Weissenborn,
  Xiaohua Zhai, Thomas Unterthiner, Mostafa Dehghani, Matthias Minderer, Georg
  Heigold, Sylvain Gelly, et~al. 2020.
\newblock An image is worth 16x16 words: Transformers for image recognition at
  scale.
\newblock \emph{arXiv preprint arXiv:2010.11929}.

\bibitem[{Evci et~al.(2022)Evci, Vladymyrov, Unterthiner, van Merri{\"e}nboer,
  and Pedregosa}]{evci2022gradmax}
Utku Evci, Max Vladymyrov, Thomas Unterthiner, Bart van Merri{\"e}nboer, and
  Fabian Pedregosa. 2022.
\newblock Gradmax: Growing neural networks using gradient information.
\newblock \emph{arXiv preprint arXiv:2201.05125}.

\bibitem[{Gardent et~al.(2017)Gardent, Shimorina, Narayan, and
  Perez-Beltrachini}]{gardent2017webnlg}
Claire Gardent, Anastasia Shimorina, Shashi Narayan, and Laura
  Perez-Beltrachini. 2017.
\newblock The webnlg challenge: Generating text from rdf data.
\newblock In \emph{Proceedings of the 10th International Conference on Natural
  Language Generation}, pages 124--133.

\bibitem[{He et~al.(2021)He, Zhou, Ma, Berg-Kirkpatrick, and
  Neubig}]{he2021towards}
Junxian He, Chunting Zhou, Xuezhe Ma, Taylor Berg-Kirkpatrick, and Graham
  Neubig. 2021.
\newblock Towards a unified view of parameter-efficient transfer learning.
\newblock \emph{arXiv preprint arXiv:2110.04366}.

\bibitem[{He et~al.(2016)He, Zhang, Ren, and Sun}]{he2016deep}
Kaiming He, Xiangyu Zhang, Shaoqing Ren, and Jian Sun. 2016.
\newblock Deep residual learning for image recognition.
\newblock In \emph{Proceedings of the IEEE conference on computer vision and
  pattern recognition}, pages 770--778.

\bibitem[{Hinton et~al.(2015)Hinton, Vinyals, Dean
  et~al.}]{hinton2015distilling}
Geoffrey Hinton, Oriol Vinyals, Jeff Dean, et~al. 2015.
\newblock Distilling the knowledge in a neural network.
\newblock \emph{arXiv preprint arXiv:1503.02531}, 2(7).

\bibitem[{Hinton et~al.(2012)Hinton, Srivastava, Krizhevsky, Sutskever, and
  Salakhutdinov}]{hinton2012improving}
Geoffrey~E Hinton, Nitish Srivastava, Alex Krizhevsky, Ilya Sutskever, and
  Ruslan~R Salakhutdinov. 2012.
\newblock Improving neural networks by preventing co-adaptation of feature
  detectors.
\newblock \emph{arXiv preprint arXiv:1207.0580}.

\bibitem[{Hou et~al.(2020)Hou, Huang, Shang, Jiang, Chen, and
  Liu}]{hou2020dynabert}
Lu~Hou, Zhiqi Huang, Lifeng Shang, Xin Jiang, Xiao Chen, and Qun Liu. 2020.
\newblock Dynabert: Dynamic bert with adaptive width and depth.
\newblock \emph{Advances in Neural Information Processing Systems},
  33:9782--9793.

\bibitem[{Houlsby et~al.(2019{\natexlab{a}})Houlsby, Giurgiu, Jastrzebski,
  Morrone, De~Laroussilhe, Gesmundo, Attariyan, and
  Gelly}]{houlsby2019parameter}
Neil Houlsby, Andrei Giurgiu, Stanislaw Jastrzebski, Bruna Morrone, Quentin
  De~Laroussilhe, Andrea Gesmundo, Mona Attariyan, and Sylvain Gelly.
  2019{\natexlab{a}}.
\newblock Parameter-efficient transfer learning for nlp.
\newblock In \emph{International Conference on Machine Learning}, pages
  2790--2799. PMLR.

\bibitem[{Houlsby et~al.(2019{\natexlab{b}})Houlsby, Giurgiu, Jastrzebski,
  Morrone, De~Laroussilhe, Gesmundo, Attariyan, and Gelly}]{adapter}
Neil Houlsby, Andrei Giurgiu, Stanislaw Jastrzebski, Bruna Morrone, Quentin
  De~Laroussilhe, Andrea Gesmundo, Mona Attariyan, and Sylvain Gelly.
  2019{\natexlab{b}}.
\newblock Parameter-efficient transfer learning for nlp.
\newblock In \emph{International Conference on Machine Learning}, pages
  2790--2799. PMLR.

\bibitem[{Howard et~al.(2019)Howard, Sandler, Chu, Chen, Chen, Tan, Wang, Zhu,
  Pang, Vasudevan et~al.}]{howard2019searching}
Andrew Howard, Mark Sandler, Grace Chu, Liang-Chieh Chen, Bo~Chen, Mingxing
  Tan, Weijun Wang, Yukun Zhu, Ruoming Pang, Vijay Vasudevan, et~al. 2019.
\newblock Searching for mobilenetv3.
\newblock In \emph{Proceedings of the IEEE/CVF international conference on
  computer vision}, pages 1314--1324.

\bibitem[{Hu et~al.(2021{\natexlab{a}})Hu, Shen, Wallis, Allen-Zhu, Li, Wang,
  Wang, and Chen}]{hu2021lora}
Edward~J Hu, Yelong Shen, Phillip Wallis, Zeyuan Allen-Zhu, Yuanzhi Li, Shean
  Wang, Lu~Wang, and Weizhu Chen. 2021{\natexlab{a}}.
\newblock Lora: Low-rank adaptation of large language models.
\newblock \emph{arXiv preprint arXiv:2106.09685}.

\bibitem[{Hu et~al.(2021{\natexlab{b}})Hu, Shen, Wallis, Allen-Zhu, Li, Wang,
  Wang, and Chen}]{lora}
Edward~J Hu, Yelong Shen, Phillip Wallis, Zeyuan Allen-Zhu, Yuanzhi Li, Shean
  Wang, Lu~Wang, and Weizhu Chen. 2021{\natexlab{b}}.
\newblock Lora: Low-rank adaptation of large language models.
\newblock \emph{arXiv preprint arXiv:2106.09685}.

\bibitem[{Jafari et~al.(2021)Jafari, Rezagholizadeh, Sharma, and
  Ghodsi}]{jafari2021annealing}
Aref Jafari, Mehdi Rezagholizadeh, Pranav Sharma, and Ali Ghodsi. 2021.
\newblock \href {https://aclanthology.org/2021.eacl-main.212} {Annealing
  knowledge distillation}.
\newblock In \emph{Proceedings of the 16th Conference of the European Chapter
  of the Association for Computational Linguistics: Main Volume}, pages
  2493--2504, Online. Association for Computational Linguistics.

\bibitem[{Jiao et~al.(2019)Jiao, Yin, Shang, Jiang, Chen, Li, Wang, and
  Liu}]{jiao2019tinybert}
Xiaoqi Jiao, Yichun Yin, Lifeng Shang, Xin Jiang, Xiao Chen, Linlin Li, Fang
  Wang, and Qun Liu. 2019.
\newblock Tinybert: Distilling bert for natural language understanding.
\newblock \emph{arXiv preprint arXiv:1909.10351}.

\bibitem[{Karimi~Mahabadi et~al.(2021)Karimi~Mahabadi, Henderson, and
  Ruder}]{karimi2021compacter}
Rabeeh Karimi~Mahabadi, James Henderson, and Sebastian Ruder. 2021.
\newblock Compacter: Efficient low-rank hypercomplex adapter layers.
\newblock \emph{Advances in Neural Information Processing Systems},
  34:1022--1035.

\bibitem[{Lester et~al.(2021)Lester, Al-Rfou, and Constant}]{lester2021power}
Brian Lester, Rami Al-Rfou, and Noah Constant. 2021.
\newblock The power of scale for parameter-efficient prompt tuning.
\newblock \emph{arXiv preprint arXiv:2104.08691}.

\bibitem[{Li et~al.(2021)Li, Lin, Ren, Li, Zhou, and Sun}]{li2021dynamic}
Lei Li, Yankai Lin, Shuhuai Ren, Peng Li, Jie Zhou, and Xu~Sun. 2021.
\newblock Dynamic knowledge distillation for pre-trained language models.
\newblock \emph{arXiv preprint arXiv:2109.11295}.

\bibitem[{Li et~al.(2019)Li, Yatskar, Yin, Hsieh, and Chang}]{li2019visualbert}
Liunian~Harold Li, Mark Yatskar, Da~Yin, Cho-Jui Hsieh, and Kai-Wei Chang.
  2019.
\newblock Visualbert: A simple and performant baseline for vision and language.
\newblock \emph{arXiv preprint arXiv:1908.03557}.

\bibitem[{Liu et~al.(2019)Liu, Ott, Goyal, Du, Joshi, Chen, Levy, Lewis,
  Zettlemoyer, and Stoyanov}]{liu2019roberta}
Yinhan Liu, Myle Ott, Naman Goyal, Jingfei Du, Mandar Joshi, Danqi Chen, Omer
  Levy, Mike Lewis, Luke Zettlemoyer, and Veselin Stoyanov. 2019.
\newblock Roberta: A robustly optimized bert pretraining approach.
\newblock \emph{arXiv preprint arXiv:1907.11692}.

\bibitem[{Lu et~al.(2019)Lu, Batra, Parikh, and Lee}]{lu2019vilbert}
Jiasen Lu, Dhruv Batra, Devi Parikh, and Stefan Lee. 2019.
\newblock Vilbert: Pretraining task-agnostic visiolinguistic representations
  for vision-and-language tasks.
\newblock \emph{Advances in neural information processing systems}, 32.

\bibitem[{Mao et~al.(2021)Mao, Mathias, Hou, Almahairi, Ma, Han, Yih, and
  Khabsa}]{mao2021unipelt}
Yuning Mao, Lambert Mathias, Rui Hou, Amjad Almahairi, Hao Ma, Jiawei Han,
  Wen-tau Yih, and Madian Khabsa. 2021.
\newblock Unipelt: A unified framework for parameter-efficient language model
  tuning.
\newblock \emph{arXiv preprint arXiv:2110.07577}.

\bibitem[{Nan et~al.(2020)Nan, Radev, Zhang, Rau, Sivaprasad, Hsieh, Tang,
  Vyas, Verma, Krishna et~al.}]{nan2020dart}
Linyong Nan, Dragomir Radev, Rui Zhang, Amrit Rau, Abhinand Sivaprasad,
  Chiachun Hsieh, Xiangru Tang, Aadit Vyas, Neha Verma, Pranav Krishna, et~al.
  2020.
\newblock Dart: Open-domain structured data record to text generation.
\newblock \emph{arXiv preprint arXiv:2007.02871}.

\bibitem[{Noach and Goldberg(2020)}]{noach2020compressing}
Matan~Ben Noach and Yoav Goldberg. 2020.
\newblock Compressing pre-trained language models by matrix decomposition.
\newblock In \emph{Proceedings of the 1st Conference of the Asia-Pacific
  Chapter of the Association for Computational Linguistics and the 10th
  International Joint Conference on Natural Language Processing}, pages
  884--889.

\bibitem[{Novikova et~al.(2017)Novikova, Du{\v{s}}ek, and
  Rieser}]{novikova2017e2e}
Jekaterina Novikova, Ond{\v{r}}ej Du{\v{s}}ek, and Verena Rieser. 2017.
\newblock The e2e dataset: New challenges for end-to-end generation.
\newblock \emph{arXiv preprint arXiv:1706.09254}.

\bibitem[{Passban et~al.(2021)Passban, Wu, Rezagholizadeh, and
  Liu}]{passban2020alp}
Peyman Passban, Yimeng Wu, Mehdi Rezagholizadeh, and Qun Liu. 2021.
\newblock \href {https://ojs.aaai.org/index.php/AAAI/article/view/17610}
  {{ALP-KD:} attention-based layer projection for knowledge distillation}.
\newblock In \emph{Thirty-Fifth {AAAI} Conference on Artificial Intelligence,
  {AAAI} 2021, Thirty-Third Conference on Innovative Applications of Artificial
  Intelligence, {IAAI} 2021, The Eleventh Symposium on Educational Advances in
  Artificial Intelligence, {EAAI} 2021, Virtual Event, February 2-9, 2021},
  pages 13657--13665. {AAAI} Press.

\bibitem[{Prato et~al.(2020)Prato, Charlaix, and
  Rezagholizadeh}]{prato2020fully}
Gabriele Prato, Ella Charlaix, and Mehdi Rezagholizadeh. 2020.
\newblock Fully quantized transformer for machine translation.
\newblock In \emph{Findings of the Association for Computational Linguistics:
  EMNLP 2020}, pages 1--14.

\bibitem[{Rashid et~al.(2021)Rashid, Lioutas, and
  Rezagholizadeh}]{rashid2021mate}
Ahmad Rashid, Vasileios Lioutas, and Mehdi Rezagholizadeh. 2021.
\newblock Mate-kd: Masked adversarial text, a companion to knowledge
  distillation.
\newblock \emph{arXiv preprint arXiv:2105.05912}.

\bibitem[{Rippel et~al.(2014)Rippel, Gelbart, and Adams}]{nested_drop}
Oren Rippel, Michael Gelbart, and Ryan Adams. 2014.
\newblock \href {https://proceedings.mlr.press/v32/rippel14.html} {Learning
  ordered representations with nested dropout}.
\newblock In \emph{Proceedings of the 31st International Conference on Machine
  Learning}, volume~32 of \emph{Proceedings of Machine Learning Research},
  pages 1746--1754, Bejing, China. PMLR.

\bibitem[{Simonyan and Zisserman(2014)}]{simonyan2014very}
Karen Simonyan and Andrew Zisserman. 2014.
\newblock Very deep convolutional networks for large-scale image recognition.
\newblock \emph{arXiv preprint arXiv:1409.1556}.

\bibitem[{Su et~al.(2019)Su, Zhu, Cao, Li, Lu, Wei, and Dai}]{su2019vl}
Weijie Su, Xizhou Zhu, Yue Cao, Bin Li, Lewei Lu, Furu Wei, and Jifeng Dai.
  2019.
\newblock Vl-bert: Pre-training of generic visual-linguistic representations.
\newblock \emph{arXiv preprint arXiv:1908.08530}.

\bibitem[{Sun et~al.(2022)Sun, Shao, Qian, Huang, and Qiu}]{sun2022black}
Tianxiang Sun, Yunfan Shao, Hong Qian, Xuanjing Huang, and Xipeng Qiu. 2022.
\newblock Black-box tuning for language-model-as-a-service.
\newblock \emph{arXiv preprint arXiv:2201.03514}.

\bibitem[{Tahaei et~al.(2021)Tahaei, Charlaix, Nia, Ghodsi, and
  Rezagholizadeh}]{tahaei2021kroneckerbert}
Marzieh~S Tahaei, Ella Charlaix, Vahid~Partovi Nia, Ali Ghodsi, and Mehdi
  Rezagholizadeh. 2021.
\newblock Kroneckerbert: Learning kronecker decomposition for pre-trained
  language models via knowledge distillation.
\newblock \emph{arXiv preprint arXiv:2109.06243}.

\bibitem[{Tao et~al.(2022)Tao, Hou, Zhang, Shang, Jiang, Liu, Luo, and
  Wong}]{tao2022compression}
Chaofan Tao, Lu~Hou, Wei Zhang, Lifeng Shang, Xin Jiang, Qun Liu, Ping Luo, and
  Ngai Wong. 2022.
\newblock Compression of generative pre-trained language models via
  quantization.
\newblock \emph{arXiv preprint arXiv:2203.10705}.

\bibitem[{Wang et~al.(2020)Wang, Tang, Duan, Wei, Huang, Cao, Jiang, Zhou
  et~al.}]{wang2020k}
Ruize Wang, Duyu Tang, Nan Duan, Zhongyu Wei, Xuanjing Huang, Guihong Cao,
  Daxin Jiang, Ming Zhou, et~al. 2020.
\newblock K-adapter: Infusing knowledge into pre-trained models with adapters.
\newblock \emph{arXiv preprint arXiv:2002.01808}.

\bibitem[{Wang et~al.(2019)Wang, Wohlwend, and Lei}]{wang2019structured}
Ziheng Wang, Jeremy Wohlwend, and Tao Lei. 2019.
\newblock Structured pruning of large language models.
\newblock \emph{arXiv preprint arXiv:1910.04732}.

\bibitem[{Xia et~al.(2021)Xia, Huang, Duan, Zhang, Ji, Sui, Cui, Bharti, and
  Zhou}]{xia2021xgpt}
Qiaolin Xia, Haoyang Huang, Nan Duan, Dongdong Zhang, Lei Ji, Zhifang Sui,
  Edward Cui, Taroon Bharti, and Ming Zhou. 2021.
\newblock Xgpt: Cross-modal generative pre-training for image captioning.
\newblock In \emph{CCF International Conference on Natural Language Processing
  and Chinese Computing}, pages 786--797. Springer.

\end{thebibliography}
\bibliographystyle{acl_natbib}

\appendix

\section{Time complexity}
\label{ap:time_complexity}

The training time for DyLoRA is comparable to that of LoRA trained once on a specific rank. Thus, when searching the rank space for LoRA, we need to train it multiple times, whereas our method does not require searching the ranks. Accordingly, DyLoRA's relative time complexity is inversely proportional to the number of possible ranks for which the LoRA model must be searched. In MRPC, DyLoRA (for all the ranks) and LoRA (only on a single rank 8) require a total training time of 408.39 seconds and 399.95 seconds, respectively. Consequently, when we need to train eight LoRA models (Rank=1,2,...,8), it will result in a cost of 399.95*8=3199.6s, compared to the training time of our model, which is only 408.39 seconds.





\section{Hyperparameters}
\label{ap:hyperparameters}
We did not use any parameter tuning nor MNLI trick (initializing some down-streams tasks from MNLI checkpoint instead of Pre-trained weights). Therefore, we fine-tuned all the datasets from original pre-trained weights. We simply followed a unified hyper-parameters for all different experiments. Unlike LoRA \cite{hu2021lora} which reported the median over 5 random seeds, we reported the mean and standard deviation over 5 random seeds. See the details in Table \ref{hyperparamters-table}:

\begin{table*}[hbt!]
\centering
\begin{tabular}{l|cc}
\hline
\textbf{Model} & \textbf{Parameter} & \textbf{Value} \\
\hline
\multirow{12}{*}{RoBERTa-Base} & &  \\
& Optimizer & AdamW \\
& Warmup Ratio & 0.06 \\
& LR Scheduler & Linear \\
& Batch Size & 32 \\
& Epochs & 30 \\
& Learning Rate (LR) & 4e-4 \\
& Weight Decay & 0.1 \\
& LoRA Config & $r_q=r_v=8$ (unless otherwise mentioned) \\
& LoRA $\alpha$ & 16 \\
& Max Sequence Length & 512 \\
& Seeds & 10, 42, 4242, 10, 1010\\
& GPU & Tesla V100-PCIE-32GB \\
\hline
\multirow{17}{*}{GPT Medium} & &  \\
& Optimizer & AdamW \\
& Adam Beta2 & 0.999 \\
& Warmup Steps & 500 \\
& Clip & 0.0 \\
& LR Scheduler & Linear \\
& Batch Size & 8 \\
& Epochs & 5 \\
& Learning Rate (LR) & 2e-4 \\
& Weight Decay & 0.01 \\
& Correct Bias & True \\
& LoRA Dropout & 0.1 \\
& Lable Smooth & 0.1 \\
& LoRA Config & $r_q=r_v=4$ \\
& LoRA $\alpha$ & 32 \\
& Seeds & 10, 42, 4242\\
& GPU & Tesla V100-PCIE-32GB \\
\hline
\end{tabular}
\caption{
The hyperparameters that has been used in our experiment for all the experiments.
}
\label{hyperparamters-table}
\end{table*}

\section{Algorithm}
\label{ap:algorithm}

A description of the proposed method is available in Algorithm \ref{alg:two}.

\begin{algorithm}[htb!]
\caption{DyLoRA - Training}\label{alg:two}
\begin{algorithmic}
\REQUIRE \\
$r\in$Range[$r_{min}$,$r_{max}$]; $i$: the number of training iterations; $\alpha$: a scaling factor; $p_B$: probability distribution function for rank selection; $X \in \mathbb{R}^{d \times n}$ : all input features to LORA; $W_0 \in \mathbb{R}^{m \times d}$ the original frozen pre-trained weight matrix
\REQUIRE $W_{dw} \in \mathbb{R}^{r \times d}$; $W_{up} \in \mathbb{R}^{m \times r}$, FROZEN: whether to keep the lower ranks frozen when updating the higher ranks
\WHILE {t < $i$}:
    \STATE {\color{blue} Forward:}
    \STATE {\color{gray} // sample a specific rank, during test is given}
    \STATE $b\sim p_B(.)$
    \STATE {\color{gray} // truncate down-projection matrix}
    \STATE $W_{dw \downarrow b} = W_{dw}$[:$b$,:]
    \STATE $W_{dw}^b = W_{dw}$[$b$,:]
    \STATE {\color{gray} // truncate up-projection matrix}
    \STATE $W_{up \downarrow b} = W_{up}$[:,:$b$]
    \STATE $W_{up}^{b} = W_{up}$[:,$b$]
    \STATE {\color{gray} // calculate the LoRA output}
    \STATE $h = W_0 X + \frac{\alpha}{b} W_{up\downarrow b} W_{dw\downarrow b}X$
    
    \STATE {\color{blue} Backward:}
    \IF{FROZEN}
    \STATE {\color{gray} \hspace{20pt} // only update the unique parameters of the selected rank}
    \STATE \hspace{20pt} $W_{dw}^b  \leftarrow W_{dw}^b - \eta \nabla_{W_{dw}^b} \mathcal{L}_{\downarrow b}^{\mathcal{DY}}$
    
    \STATE \hspace{20pt} $W_{up}^b  \leftarrow W_{up}^b - \eta \nabla_{W_{up}^b} \mathcal{L}_{\downarrow b}^{\mathcal{DY}}$
    \ELSE
    \STATE \hspace{20pt} $W_{dw\downarrow b}  \leftarrow W_{dw\downarrow b} - \eta \nabla_{W_{dw\downarrow b}^b} \mathcal{L}_{\downarrow b}^{\mathcal{DY}}$
    
    \STATE \hspace{20pt} $W_{up\downarrow b}  \leftarrow W_{up\downarrow b} - \eta \nabla_{W_{up\downarrow b}^b} \mathcal{L}_{\downarrow b}^{\mathcal{DY}}$
    \ENDIF
\ENDWHILE

\end{algorithmic}
\label{alg}
\end{algorithm}



\section{GPT Experiments}
\label{ap:gpt-experiments}

A summary of the additional experiments that have been conducted to demonstrate the effectiveness of our proposed method for the task of language generation is provided in Table \ref{gpt-experiments-e2e-appendix} and Table \ref{gpt-experiments-appendix-Dart-webnlg-appendix}.

\begin{table*}[hbt!]
\centering
\resizebox{\textwidth}{!}{
\begin{tabular}{lcc|cccccc}
\hline
\textbf{Model} (Method) & \textbf{Updated Params} & \textbf{Trainable Params} & \multicolumn{5}{c}{E2E NLG Challenge} \\ 
& & & \textbf{BLEU} & \textbf{NIST} & \textbf{MET} & \textbf{ROUGE-L} & \textbf{CIDEr} \\
\hline
\multicolumn{8}{c}{Rank=1} \\
\hline
GPT-2 M (LoRA) & & 0.09M & 3.4 & 1.2 & 9.2 & 18.8 & 0.1 \\
\textbf{GPT-2 M (DyLoRA)} & $W_{dw}^b$,$W_{up}^b$ & 0.09M & $67.9_{\pm0.2}$ & $8.6_{\pm0.1}$ & $44.9_{\pm0.4}$ & $69.1_{\pm0.3}$  & $2.4_{\pm0.0}$ \\
\hline
\multicolumn{8}{c}{Rank=2} \\
\hline
GPT-2 M (LoRA) & & 0.19M & 47.0 & 6.4 & 34.2 & 56.1 & 1.3 \\
\textbf{GPT-2 M (DyLoRA)} & $W_{dw}^b$,$W_{up}^b$  & 0.19M & $68.8_{\pm0.5}$ & $8.7_{\pm0.02}$ & $45.2_{\pm0.2}$ & $69.8_{\pm0.3}$  & $2.4_{\pm0.0}$ \\
\hline
\multicolumn{8}{c}{Rank=3} \\
\hline
GPT-2 M (LoRA) & & 0.29M & 63.7 & 8.5 & 42.4 & 65.8 & 2.2 \\
\textbf{GPT-2 M (DyLoRA)} & $W_{dw}^b$,$W_{up}^b$  & 0.29M & $68.4_{\pm1.0}$ & $8.7_{\pm0.1}$ & $45.3_{\pm0.6}$ & $69.8_{\pm0.7}$  & $2.4_{\pm0.0}$ \\
\hline
\multicolumn{8}{c}{Rank=4} \\
\hline
GPT-2 M (LoRA) & & 0.39M & 69.9 & 8.8 & 46.8 & 72.1 & 2.5 \\
\textbf{GPT-2 M (DyLoRA)} & $W_{dw}^b$,$W_{up}^b$  & 0.39M & $68.4_{\pm0.41}$ & $8.7_{\pm0.02}$ & $45.5_{\pm0.6}$ & $69.9_{\pm0.5}$  & $2.4_{\pm0.0}$ \\
\hline
\multicolumn{8}{c}{Fine-Tune} \\
\hline
GPT-2 M (FT)$^*$ & & 354M & 68.2 & 8.62 & 46.2 & 71.0 & 2.5  \\
\hline
\end{tabular}
}
\caption{
For all metrics, higher is better. Rows with * have been reported based on the LoRA paper. Unlike \cite{hu2021lora}, we included the classifier number of parameters in our trainable parameters count. 
}
\label{gpt-experiments-e2e-appendix}
\end{table*}

\begin{table*}[hbt!]
\centering
\resizebox{\textwidth}{!}{
\begin{tabular}{lcccccc}
\hline
\textbf{Model} (Method) & \textbf{Trainable Params} & \multicolumn{2}{c}{DART} & \multicolumn{2}{c}{WebNLG}\\ 
& & \textbf{BLEU$\uparrow$} & \textbf{TER$\downarrow$} & \textbf{BLEU$\uparrow$} & \textbf{TER$\downarrow$}\\
\hline
\multicolumn{4}{c}{Rank=1} \\ 
\hline
GPT-2 M (LoRA) & 0.09M & 0.7 & 0.49 & 2.8 & 1.2 \\
GPT-2 M (DyLoRA-Frozen) & 0.09M & $44.5_{\pm0.1}$ & $0.5_{\pm0.0}$& $52.1_{\pm0.0}$ & $0.4_{\pm0.0}$ \\
\hline
\multicolumn{4}{c}{Rank=2} \\ 
\hline
GPT-2 M (LoRA) & 0.19M & 15.9 & 0.48 & 26.6 & 0.67 \\
GPT-2 M (DyLoRA-Frozen) & 0.19M & $45.0_{\pm0.1}$ & $0.5_{\pm0.0}$& $52.7_{\pm0.3}$ & $0.4_{\pm0.0}$ \\
\hline
\multicolumn{4}{c}{Rank=3} \\ 
\hline
GPT-2 M (LoRA) & 0.29M & 35.8 & 0.47& 43.6 & 0.47 \\
GPT-2 M (DyLoRA-Frozen) & 0.29M & $45.2_{\pm0.1}$ & $0.5_{\pm0.0}$ & $53.0_{\pm0.5}$ & $0.4_{\pm0.0}$\\
\hline
\multicolumn{4}{c}{Rank=4} \\ 
\hline
GPT-2 M (LoRA) & 0.39M & 48.0 & 0.47 & 55.57 & 0.39 \\
GPT-2 M (DyLoRA-Frozen) & 0.39M & $45.6_{\pm0.3}$ & $0.5_{\pm0.0}$ & $53.0_{\pm0.0}$ & $0.4_{\pm0.0}$ \\
\hline
\multicolumn{4}{c}{Fine-Tune} \\ 
\hline
GPT-2 M (FT)$^*$ & 354M & 46.2 & 0.46 &  &  \\
\hline
\end{tabular}
}
\caption{
Rows with * have been reported based on the LoRA paper. \cite{hu2021lora}. In $FT^{Top2}$, only the last two layers have been fine-tuned.
}
\label{gpt-experiments-appendix-Dart-webnlg-appendix}
\end{table*}


\section{RoBERTa-Base Experiments}
\label{ap:roberta-base-experiments}

Additional experiments for RoBERTa-Base that we were unable to include in the main content may be accessed in Table \ref{fixed-budget-frozen-appendix}. In the attachment, we have included all the details and results for the majority of the experiments, in addition to the code.

\begin{table*}[hbt!]
\centering
\resizebox{\textwidth}{!}{
\begin{tabular}{lccccccccc}
 & \textbf{Accuracy} & \textbf{Accuracy} & \textbf{F1} & \textbf{Mathews} & \textbf{Accuracy} & \textbf{Accuracy} & \textbf{Accuracy} & \textbf{Pearson} & \\
\hline
\textbf{Model} & \textbf{MNLI} & \textbf{SST-2} & \textbf{MRPC} & \textbf{CoLA} & \textbf{QNLI} & \textbf{QQP} & \textbf{RTE} & \textbf{STS-B} & \textbf{Avg}\\
\hline
\multicolumn{10}{c}{Rank = 1} \\ 
\hline
LoRA & $34.60_{\pm3.69}$ &	$69.61_{\pm7.99}$&	$83.47_{\pm3.90}$& $25.57_{\pm9.71}$&	$53.00_{\pm2.95}$&	$44.30_{\pm7.50}$&	$57.55_{\pm5.51}$&	$76.07_{\pm6.06}$&	54.90 \\
\textbf{DyLoRA (Frozen)} & $85.36_{\pm0.26}$ &	$93.51_{\pm0.49}$&	$90.75_{\pm0.70}$& $56.95_{\pm1.54}$&	$91.70_{\pm0.28}$&	$87.87_{\pm0.17}$&	$66.79_{\pm8.54}$&	$89.95_{\pm0.24}$&	82.86 \\
\textbf{DyLoRA} & $85.59_{\pm0.07}$ &	$93.23_{\pm0.63}$&	$91.58_{\pm0.69}$& $57.93_{\pm2.12}$&	$91.95_{\pm0.14}$&	$88.37_{\pm0.15}$&	$74.80_{\pm1.48}$&	$90.30_{\pm0.13}$&	\textbf{84.22} \\
\hline
\multicolumn{10}{c}{Rank = 2} \\ 
\hline
LoRA &  $40.53_{\pm6.17}$ &	$82.75_{\pm5.08}$&	$88.00_{\pm1.81}$& $43.30_{\pm4.67}$&	$63.42_{\pm2.99}$&	$59.21_{\pm6.13}$&	$68.88_{\pm1.26}$&	$85.51_{\pm1.94}$&	66.45 \\
\textbf{DyLoRA (Frozen)} & $85.74_{\pm0.28}$ &	$93.76_{\pm0.52}$&	$91.09_{\pm0.45}$& $56.88_{\pm2.09}$&	$92.03_{\pm0.22}$&	$88.21_{\pm0.07}$&	$63.90_{\pm12.85}$&	$90.25_{\pm0.15}$&	82.73 \\
\textbf{DyLoRA} & $86.02_{\pm0.06}$ &	$93.81_{\pm0.30}$&	$91.66_{\pm0.46}$& $59.91_{\pm1.88}$&	$92.39_{\pm0.25}$&	$89.33_{\pm0.05}$&	$76.03_{\pm1.61}$&	$90.60_{\pm0.09}$&	\textbf{84.97} \\
\hline
\multicolumn{10}{c}{Rank = 3} \\ 
\hline
LoRA &  $58.95_{\pm6.02}$ &	$90.00_{\pm1.27}$&	$89.66_{\pm1.25}$& $56.78_{\pm1.88}$&	$79.26_{\pm4.80}$&	$72.58_{\pm4.09}$&	$72.49_{\pm2.30}$&	$88.80_{\pm0.29}$&	76.07 \\
\textbf{DyLoRA (Frozen)} & $85.78_{\pm0.25}$ &	$93.76_{\pm0.26}$&	$91.78_{\pm0.89}$& $58.86_{\pm0.32}$&	$92.17_{\pm0.18}$&	$88.40_{\pm0.0}$&	$70.90_{\pm6.14}$&	$90.50_{\pm0.29}$&	84.02 \\
\textbf{DyLoRA} & $86.70_{\pm0.09}$ &	$94.11_{\pm0.33}$&	$91.56_{\pm0.86}$& $60.97_{\pm2.01}$&	$92.77_{\pm0.21}$&	$89.76_{\pm0.07}$&	$77.11_{\pm2.97}$&	$90.69_{\pm0.14}$&	\textbf{85.46} \\
\hline
\multicolumn{10}{c}{Rank = 4} \\ 
\hline
LoRA &  $72.10_{\pm5.25}$ &	$91.56_{\pm0.34}$&	$89.62_{\pm0.92}$& $58.53_{\pm3.93}$&	$85.09_{\pm1.20}$&	$80.78_{\pm3.73}$&	$73.07_{\pm2.29}$&	$89.28_{\pm0.72}$&	80.00 \\
\textbf{DyLoRA (Frozen)} & $85.93_{\pm0.19}$ &	$93.85_{\pm0.33}$&	$91.28_{\pm0.71}$& $59.25_{\pm1.05}$&	$92.27_{\pm0.16}$&	$88.52_{\pm0.08}$&	$71.12_{\pm2.46}$&	$90.53_{\pm0.18}$&	84.10 \\
\textbf{DyLoRA} & $86.82_{\pm0.04}$ &	$94.40_{\pm0.13}$&	$92.06_{\pm0.46}$& $59.81_{\pm1.71}$&	$92.91_{\pm0.31}$&	$89.80_{\pm0.10}$&	$77.40_{\pm2.72}$&	$90.86_{\pm0.06}$&	\textbf{85.53} \\
\hline
\multicolumn{10}{c}{Rank = 5} \\ 
\hline
LoRA &  $78.61_{\pm3.97}$ &	$92.82_{\pm0.46}$&	$90.75_{\pm0.96}$& $60.37_{\pm3.10}$&	$88.97_{\pm0.90}$&	$85.26_{\pm1.56}$&	$73.21_{\pm2.17}$&	$89.90_{\pm0.30}$&	82.49 \\
\textbf{DyLoRA (Frozen)} & $85.95_{\pm0.17}$ &	$93.78_{\pm0.26}$&	$91.28_{\pm0.64}$& $59.41_{\pm1.30}$&	$92.30_{\pm0.17}$&	$88.56_{\pm0.09}$&	$71.48_{\pm2.92}$&	$90.60_{\pm0.20}$&	84.17 \\
\textbf{DyLoRA} & $87.00_{\pm0.10}$ &	$94.29_{\pm0.41}$&	$91.73_{\pm0.60}$& $60.52_{\pm1.07}$&	$93.01_{\pm0.28}$&	$90.04_{\pm0.10}$&	$76.90_{\pm2.11}$&	$90.97_{\pm0.20}$&	\textbf{85.56} \\
\hline
\multicolumn{10}{c}{Rank = 6} \\ 
\hline
LoRA &  $83.02_{\pm1.59}$ &	$93.49_{\pm0.88}$&	$91.28_{\pm0.63}$& $61.94_{\pm2.27}$&	$90.32_{\pm0.76}$&	$87.54_{\pm1.51}$&	$76.68_{\pm1.16}$&	$90.12_{\pm0.12}$&	84.30 \\
\textbf{DyLoRA (Frozen)} & $85.98_{\pm0.16}$ &	$93.76_{\pm0.46}$&	$91.12_{\pm0.43}$& $58.95_{\pm1.10}$&	$92.46_{\pm0.14}$&	$88.68_{\pm0.13}$&	$72.64_{\pm2.44}$&	$90.64_{\pm0.23}$&	84.28 \\
\textbf{DyLoRA} & $86.97_{\pm0.20}$ &	$94.27_{\pm0.37}$&	$91.44_{\pm0.64}$& $60.16_{\pm1.70}$&	$93.01_{\pm0.21}$&	$90.07_{\pm0.14}$&	$77.33_{\pm1.66}$&	$91.03_{\pm0.20}$&	\textbf{85.53} \\
\hline
\multicolumn{10}{c}{Rank = 7} \\ 
\hline
LoRA &  $85.44_{\pm0.78}$ &	$93.62_{\pm0.35}$&	$91.27_{\pm0.73}$& $62.19_{\pm2.66}$&	$91.88_{\pm0.23}$&	$89.51_{\pm0.30}$&	$75.52_{\pm1.41}$&	$90.35_{\pm0.24}$&	84.97 \\
\textbf{DyLoRA (Frozen)} & $86.08_{\pm0.14}$ & $93.97_{\pm0.17}$ &	$91.02_{\pm0.70}$ & $58.76_{\pm0.94}$ & $92.30_{\pm0.10}$ & $88.77_{\pm0.06}$ & $73.50_{\pm1.67}$ & $90.68_{\pm0.15}$ &	84.38 \\
\textbf{DyLoRA} & $86.82_{\pm0.10}$ & $94.27_{\pm0.33}$ &	$91.38_{\pm0.59}$ & $59.51_{\pm1.75}$ & $92.99_{\pm0.26}$ & $90.04_{\pm0.06}$ & $77.91_{\pm1.58}$ & $91.07_{\pm0.19}$ &	\textbf{85.50} \\

\hline
\multicolumn{10}{c}{Rank = 8} \\ 
\hline
LoRA &  $86.82_{\pm0.18}$ & $94.01_{\pm0.30}$ & $91.48_{\pm0.73}$ & $62.08_{\pm1.37}$ & $92.39_{\pm0.39}$ & $90.42_{\pm0.02}$ &	$74.51_{\pm0.41}$ & $90.48_{\pm0.24}$ &	85.27 \\
\textbf{DyLoRA (Frozen)} & $86.10_{\pm0.04}$ & $93.69_{\pm0.41}$ & $91.19_{\pm0.79}$ &	$58.52_{\pm0.95}$ & $92.47_{\pm0.18}$ &	$88.82_{\pm0.06}$ & $73.29_{\pm2.49}$ &	$90.68_{\pm0.14}$ & 84.35 \\
\textbf{DyLoRA} & $86.76_{\pm0.13}$ & $94.36_{\pm0.38}$ & $91.38_{\pm0.83}$ &	$59.51_{\pm1.84}$ & $93.00_{\pm0.32}$ &	$89.91_{\pm0.08}$ & $77.55_{\pm0.59}$ &	$91.05_{\pm0.19}$ & \textbf{85.44} \\

\hline
\multicolumn{10}{c}{Best (Rank)} \\ 
\hline
LoRA & 87.03(8) & 94.50(6) & 92.25(7) & \textbf{66.05}(7) & 92.81(8) & \textbf{90.45}(8) & 77.98(6) & 90.87(8) & 86.49 \\
\textbf{DyLoRA (Frozen)} & 86.18(7) & 94.50\textbf{(2)} & \textbf{92.93(3)} &	61.57(5) & 92.70\textbf{(6)} &	88.88(8) & 75.81(7) & 90.89\textbf{(6)} & 85.43 \\
\textbf{DyLoRA} & \textbf{87.17(6)} & \textbf{94.72}(7) & 92.79(8) &	63.32\textbf{(3)} & \textbf{93.56}(8) & 90.17\textbf{(6)} & \textbf{80.14(4)} & \textbf{91.36}(7) & \textbf{86.66} \\
\hline
\multicolumn{10}{c}{Full Rank} \\ 
\hline
Fine Tune$^*$ & \textbf{87.6} & \textbf{94.8} & 90.2 & 63.6 & 92.8 &\textbf{91.9} &78.7 &91.2 & 86.4 \\
\hline
\end{tabular}
}
\caption{\label{fixed-budget-frozen-appendix}
In this table, the task is to find a low-rank adaptation matrix that works with different ranks at inference time given a fixed budget (training time).
}
\end{table*}

\end{document}